\documentclass[10pt,twocolumn,letterpaper]{article}

\usepackage[pagenumbers]{cvpr}              

\usepackage[dvipsnames]{xcolor}

\definecolor{cvprblue}{rgb}{0.21,0.49,0.74}
\usepackage[pagebackref,breaklinks,colorlinks,citecolor=cvprblue]{hyperref}
\usepackage{multirow}
\usepackage[normalem]{ulem}
\usepackage[T1]{fontenc}
\usepackage{nicematrix}
\useunder{\uline}{\ul}{}
\usepackage{colortbl}
\usepackage{adjustbox} 
\usepackage{amssymb, bbding}
\definecolor{colorTab}{rgb}{0.95,0.90,0.92}

\def\paperID{12315} 
\def\confName{CVPR}
\def\confYear{2025}

\title{Noise Calibration and Spatial-Frequency Interactive Network for \\ STEM Image Enhancement}

\author{
    Hesong Li$^1$ \quad Ziqi Wu$^1$ \quad Ruiwen Shao$^1$ \quad Tao Zhang$^2$ \quad Ying Fu$^{1\dagger}$ \\
    $^1$Beijing Institute of Technology \quad 
    $^2$Hangzhou Dianzi University \\
    {\tt\small \{lihesong2,3220222208,rwshao,fuying\}@bit.edu.cn \quad tzhang@hdu.edu.cn}
}

\begin{document}
\maketitle

\renewcommand{\thefootnote}{$\dagger$}
\footnotetext{Corresponding author.}

\begin{abstract}
Scanning Transmission Electron Microscopy (STEM) enables the observation of atomic arrangements at sub-angstrom resolution, allowing for atomically resolved analysis of the physical and chemical properties of materials. However, due to the effects of noise, electron beam damage, sample thickness, etc, obtaining satisfactory atomic-level images is often challenging. Enhancing STEM images can reveal clearer structural details of materials. Nonetheless, existing STEM image enhancement methods usually overlook unique features in the frequency domain, and existing datasets lack realism and generality. To resolve these issues, in this paper, we develop noise calibration, data synthesis, and enhancement methods for STEM images. We first present a STEM noise calibration method, which is used to synthesize more realistic STEM images. The parameters of background noise, scan noise, and pointwise noise are obtained by statistical analysis and fitting of real STEM images containing atoms. Then we use these parameters to develop a more general dataset that considers both regular and random atomic arrangements and includes both HAADF and BF mode images. Finally, we design a spatial-frequency interactive network for STEM image enhancement, which can explore the information in the frequency domain formed by the periodicity of atomic arrangement. Experimental results show that our data is closer to real STEM images and achieves better enhancement performances together with our network. Code will be available at \href{https://github.com/HeasonLee/SFIN}{https://github.com/HeasonLee/SFIN}.
\end{abstract}

\setlength{\abovedisplayskip}{5pt}
\setlength{\belowdisplayskip}{5pt}

\section{Introduction}
\label{sec:intro}
Scanning Transmission Electron Microscopy (STEM) enables scientific observation at the atomic scale and plays a crucial role in the discovery and study of new materials such as semiconductor materials \cite{semiconductor,semiconductor2}, new energy materials \cite{Battery1,Battery2,Battery3}, and functional materials \cite{functional,functional2}. 

Although STEM images have the advantage of atomic scale, they are susceptible to various types of noise interference \cite{abTEM, AtomSegNet,GAN}. Since prolonged exposure or increased electron beam intensity can damage samples and cannot remove background noise caused by impurities, high-quality real STEM image data cannot be collected directly.

\begin{figure}
\centering
\includegraphics[width=1\linewidth]{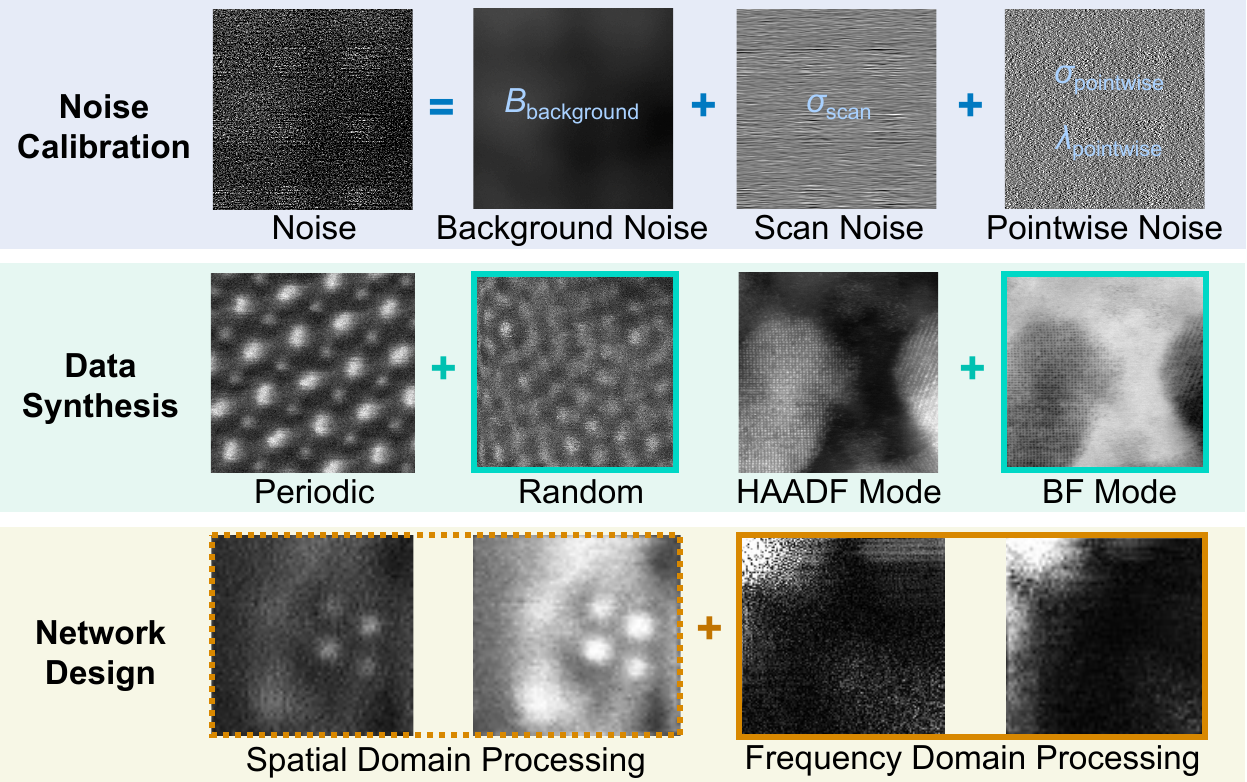}
\vspace{-6mm}
\caption{The three efforts we undertake to boost the performance of STEM image enhancement. (1) Calibrating the noise in real STEM images to make synthetic data more realistic; (2) Increasing the generalization of the dataset by adding randomized atomic arrangements and BF mode images; (3) Incorporating frequency domain operations in the image enhancement network to leverage the regularities in atomic arrangements.}
\vspace{-5mm}
\label{fig:head}
\end{figure}

Some existing studies have synthesized simulated STEM datasets \cite{abTEM,AtomSegNet,GAN} and trained deep neural networks \cite{AtomSegNet,FCN} to achieve STEM image enhancement. However, there are three unresolved issues with existing datasets and enhancement methods. Firstly, current synthetic data \cite{abTEM,AtomSegNet,GAN} lacks realism. STEM images have unique scanning noise related to image gradients, making traditional calibration methods requiring uniform calibration frames \cite{calib1,calib2} unavailable. Thus, existing datasets use random noise parameters, which are different from those in real STEM images. Secondly, existing datasets lack generality because they only include regular atomic patterns and exclude random arrangements such as defects, embeddings, and interfaces. Besides, they only include HAADF images, missing BF mode, which is more suitable for light atoms observation \cite{bf1, bf2,bf3}. Thirdly, existing enhancement networks \cite{AtomSegNet,FCN} usually focus on the spatial domain, ignoring the strong prior formed by periodic atomic arrangements in the frequency domain.

As shown in Figure~\ref{fig:head}, to address these issues, we first establish a noise model for STEM images and calibrate the parameters of these noises through a set of real STEM images containing atomic arrangements. By statistically analyzing and linearly fitting the histograms of real images, the parameters and fluctuations of various noise types can be obtained. Next, we synthesize simulated STEM data using the calibrated parameters, which solves the problem of inconsistent noise distribution between the previous dataset and the real situation. Based on several fixed atomic patterns, the data synthesis process adds random deletion and addition of atoms, as well as completely random atomic arrangements to accommodate atomic defects, ion embeddings, and interface scenarios. Besides HAADF mode images, we also synthesize BF mode images for observing light atoms \cite{bf1,bf2,bf3}. The addition of randomization and BF mode images increases the generality of the dataset. Finally, we design a spatial-frequency interactive network for STEM image enhancement. While restoring atomic features in the spatial domain, it utilizes the strong frequency domain information brought by the periodic arrangement of atoms. Experimental results show that our data is closer to real STEM images and achieves better performance in enhancement and some other tasks together with our spatial-frequency interactive network.

In summary, our main contributions are that we
\begin{itemize}
\item Propose a STEM noise calibration method, which can be used to synthesize more realistic STEM images.
\item Build a more general dataset that contains more types of atom arrangements and shooting modes.
\item Design a spatial-frequency interactive network, which can leverage the prior of periodic atomic arrangement.
\end{itemize}

\section{Related Work}
\label{sec:rw}

In this section, we provide an overview of the related work, including STEM image synthesis and enhancement.

\vspace{2pt}\noindent\textbf{STEM image synthesis.} STEM image synthesis provides data support for deep learning-based STEM image enhancement methods. The abTEM code library \cite{abTEM} provides a noise-free STEM image rendering method that supports multiple shooting modes but is slow and requires atomic structure files. TEMImageNet dataset \cite{AtomSegNet} contains simulated STEM images in HAADF mode and label images for STEM image enhancement and atom detection tasks such as denoising, super-resolution, and segmentation. It includes several fixed shooting field angles and rotation angles and incorporates background noise, shot noise, and scan noise with random parameters. Cycle-GAN \cite{GAN} is used to learn the mapping relationship between noiseless simulated STEM images and noisy real STEM images and generate more realistic STEM noise. This method relies entirely on reference STEM images and lacks flexibility in controlling various parameters in the data synthesis program. The calibration of real image noise \cite{calib1,calib2} has been proven to be effective in synthesizing realistic noisy images and has achieved good results in natural image enhancement. However, the intensity of scanning noise contained in STEM images depends on the gradient changes in the images, so existing natural image calibration methods \cite{calib1,calib2} based on flat field frames and bias frames without gradient changes cannot be directly used to calibrate STEM image noise. In this work, we calibrate real STEM images with atoms to synthesize more realistic STEM images and build a more general dataset that contains more types of atom arrangements and shooting modes.

\vspace{2pt}\noindent\textbf{STEM image enhancement.} Reconstruction and enhancement \cite{FCDFusion,CJE1,CJE2,CJE3,CJE4} of natural images are widely studied and applied. For STEM images, two commonly used traditional enhancement methods are Wiener filtering \cite{Wiener} and bilateral filtering \cite{Bilateral}, which have been integrated into electron microscopy data processing software such as Gatan Microscopy Suite. These methods can filter out high-frequency noise, but cannot remove smooth background noise. Recently, some deep learning-based STEM image processing methods \cite{AtomSegNet,FCN,GAN2} have emerged, which can finish various tasks with different training labels, including background removal, denoising, super-resolution, atomic segmentation, and atomic localization, etc. A fully convolutional network \cite{FCN} is used to identify atomic defects, including atomic deletions and replacements. Generative adversarial networks \cite{GAN2} are used for pre-training on large-scale cell electron microscopy dataset CEM500k \cite{CEM500k} and fine-tuning on downstream tasks such as STEM image enhancement with TEMImageNet \cite{AtomSegNet} dataset. AtomSegNet \cite{AtomSegNet} is a convolutional neural network based on UNet architecture \cite{UNet}, which is trained on TEMImageNet \cite{AtomSegNet} and used for STEM image enhancement and atomic detection. AtomSegNet plays a key role in studies on new battery materials \cite{Battery1,Battery2,Battery3}. Existing STEM image enhancement methods \cite{AtomSegNet,FCN,GAN2} usually perform only in the spatial domain. In this work, we design a spatial-frequency interactive network to utilize more information in the frequency domain.

\section{Our Method}
\label{sec:data}

In this section, we first introduce our motivation, followed by the proposed noise calibration and STEM image data synthesis methods. Finally, we present our spatial-frequency interactive network.

\subsection{Motivation}
\label{sec:moti}
STEM image enhancement aims to remove noise and background. We achieve this by noise calibration, data synthesis, and network design.

Noise calibration aims to generate more realistic STEM images. The parameters of scanning noise are related to the gradient of the image, so existing noise calibration methods based on capturing flat field frames and bias frames without gradient changes \cite{calib1,calib2} cannot be used for STEM images. In this work, we calibrate various types of noise using STEM images that include gradient variations (\emph{i.e.}, contain atoms). By statistically analyzing and linearly fitting the histograms of real images, the fluctuations of scan noise following the magnitude of gradients can be obtained. Similarly, parameters for background noise and pointwise noise can be obtained.

In data synthesis, we incorporate more random atomic arrangements and synthesize additional imaging modes to ensure the dataset can be applied to a wider range of scenarios. Random deletion and addition of atoms, as well as completely random atomic arrangements can simulate atomic defects, ion embeddings, and structure interfaces, respectively. Additionally generated BF imaging mode data can help observe light atoms \cite{bf1,bf2,bf3}.

\begin{figure}
\centering
\includegraphics[width=1\linewidth]{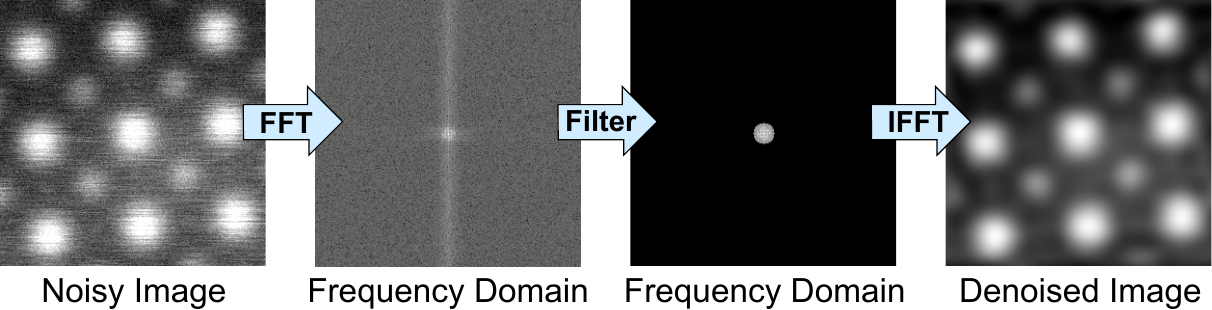}
\vspace{-7mm}
\caption{Our motivation for using frequency domain in STEM image enhancement. When atoms are arranged periodically, simple filtering of the frequency domain image corresponding to the original image can achieve a good denoising effect.}
\vspace{-3mm}
\label{fig:fft}
\end{figure}

Considering that periodic signals contain strong features in the frequency domain, we incorporate frequency domain processing to leverage the periodic information of atom arrangement. As shown in Figure~\ref{fig:fft}, filtering out important patterns in the frequency domain can help reduce noise in the spatial domain. Thus, we add frequency-domain convolutions to our STEM image enhancement network. Note that due to special circumstances such as atomic defects, ion embeddings, and interfaces, pure frequency domain information cannot represent the absence or addition of single atoms at different spatial positions. Therefore, it is necessary to combine spatial and frequency domain information to get a better enhancement result. In this work, we design a spatial-frequency interactive network for STEM image enhancement. While restoring each atomic feature in the spatial domain, it utilizes the strong frequency domain information brought by the periodic arrangement of atoms.

\subsection{STEM Noise Calibration}
\label{sec:calib}
A STEM image with noise can be represented as the sum of a clean image and various types of noise, \emph{i.e.},
\begin{equation}
\boldsymbol{I}_\text{noise}=\boldsymbol{I}_\text{clean}+ \boldsymbol{N}_\text{background}+\boldsymbol{N}_\text{scan}+\boldsymbol{N}_\text{pointwise},
\end{equation}
where $\boldsymbol{I}_\text{noise}$ and $\boldsymbol{I}_\text{clean}$ are the noisy image and the ideal clean image, respectively, while $\boldsymbol{N}_\text{background}$, $\boldsymbol{N}_\text{scan}$, and $\boldsymbol{N}_\text{pointwise}$ are the background noise, scan noise, and pointwise noise that need to be removed, respectively.

\begin{figure}
\centering
\includegraphics[width=1.\linewidth]{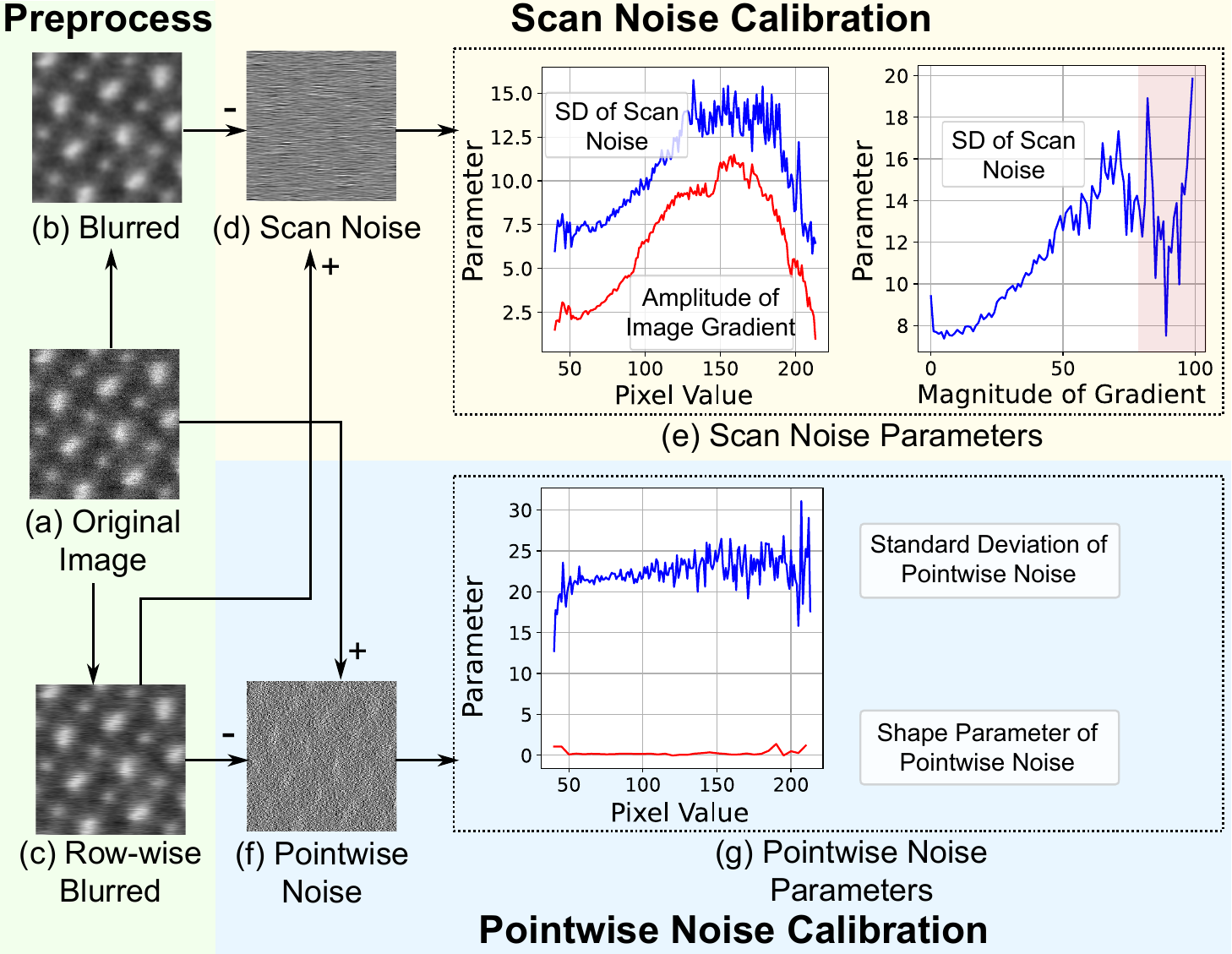}
\vspace{-7mm}
\caption{The process of calibrating scan noise and pointwise noise in our STEM noise calibration method. The light red rectangular area in subfigure (e) indicates fluctuations caused by limited data, which will be discarded during fitting.}
\vspace{-5mm}
\label{fig:calib}
\end{figure}

 \vspace{5pt}\noindent\textbf{Modeling and calibration of clean image.} 
 Theoretically, a clean STEM image consists of numerous circular spots, each corresponding to an atomic column, which can be fitted using a two-dimensional normal distribution. The type of atoms and the thickness of the atomic column affect the size and brightness of the spots. HAADF images are brighter at atomic columns due to rebounding electrons, while BF images are darker as electrons pass through \cite{bf1,bf2,bf3}. The relative brightness and size of atoms are fixed and can be simulated \cite{abTEM}. The overall brightness of a STEM image varies with capture settings, defined as the difference between the brightest center and background, 
 \begin{equation}\label{eq:b}
     b_\text{atom}=|b_\text{central}-b_\text{background}|.
 \end{equation} 
 To calibrate the brightness of a STEM image, a two-dimensional Gaussian filter is first applied to the original image to approximate noise removal, as illustrated in Figures~\ref{fig:calib}\textcolor{red}{(a)} and \ref{fig:calib}\textcolor{red}{(b)}. Then, the parameter $b_\text{atom}$ can be calculated as the difference between the maximum and minimum values of the blurred image.

 \vspace{5pt}\noindent\textbf{Modeling and calibration of background noise.} 
 Background noise $\boldsymbol{N}_\text{background}$ arises due to non-uniform sample thickness or impurities such as carbon compounds after long-term electron beam exposure. It can be simulated using Perlin noise \cite{Perlin}, which involves generating a random two-dimensional matrix where each element follows a uniform distribution, and then upsampling it to create a smoothly transitioning background noise image. Since background noise varies in value across different locations, small-sized patches (\emph{e.g.}, 256$\times$256) need to be cropped from the original image to calculate the background brightness parameter $b_\text{background}$ for each location. 

 \vspace{5pt}\noindent\textbf{Modeling and calibration of scan noise.} 
Scan noise $\boldsymbol{N}_\text{scan}$ is primarily due to the jitter of the scanning position. This results in the obtained pixel capturing random areas near the ideal imaging location. When there is significant contrast variation in these areas (such as at the edge of an atomic), the pixel value differences caused by the position shift become more pronounced. Therefore, the amplitude of scan noise is approximately proportional to the amplitude of the image gradient at that location. To calibrate its standard deviation $\sigma_\text{scan}$, it is first necessary to isolate the scan noise from the noisy STEM image. As illustrated in Figure~\ref{fig:calib}\textcolor{red}{(c)}, each row of the original image is blurred using a one-dimensional Gaussian filter, resulting in a row-wise blurred image. In it, the pointwise noise is approximately removed by the row-wise blurring, leaving only background noise and scan noise. Then, the difference between Figure~\ref{fig:calib}\textcolor{red}{(c)} and Figure~\ref{fig:calib}\textcolor{red}{(b)} represents the scan noise image, as shown in Figure~\ref{fig:calib}\textcolor{red}{(d)}. Figure~\ref{fig:calib}\textcolor{red}{(e)} presents the standard deviation of scan noise as a function of pixel brightness and image gradient amplitude, respectively. It can be observed that the standard deviation of scan noise is proportional to the amplitude of the image gradient. Thus, a linear fit can be applied to determine the slope and intercept of $\sigma_\text{scan}$, which can then be used to simulate the scan noise.

\begin{figure}
\centering
\includegraphics[width=.9\linewidth]{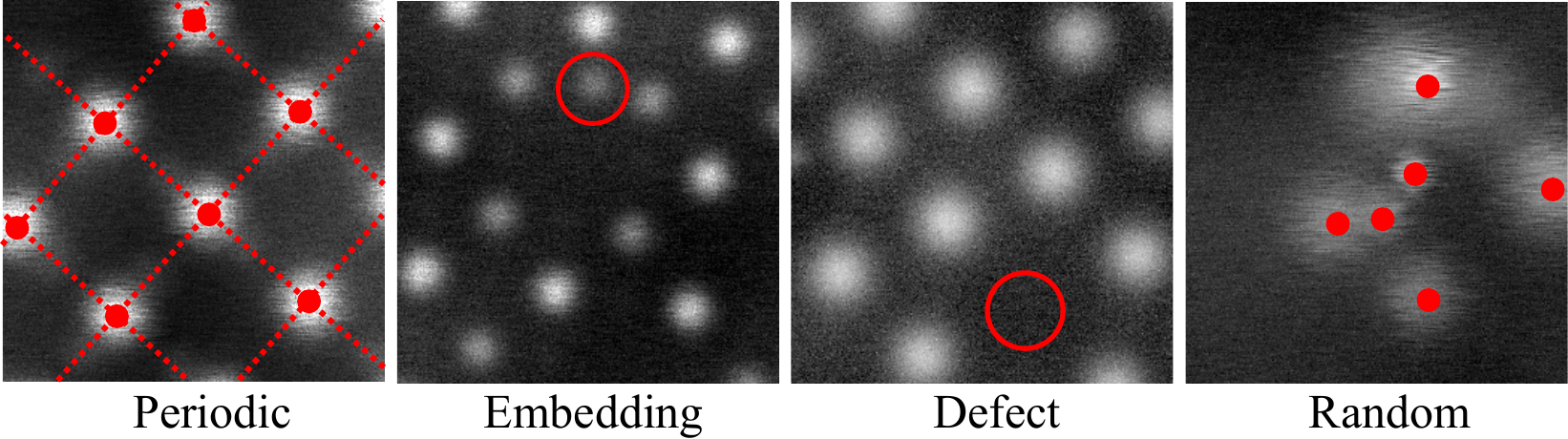}
\vspace{-3.5mm}
\caption{Our dataset simulates different atomic arrangements, including periodic patterns, atom embeddings, atom defects, and random arrangements. This enables our dataset to handle more situations. Other datasets such as GAN \cite{GAN} and TEMImageNet \cite{AtomSegNet} do not include atom embeddings and random arrangements.}
\vspace{-5mm}
\label{fig:random}
\end{figure}

 \vspace{5pt}\noindent\textbf{Modeling and calibration of pointwise noise.} 
 Pointwise noise $\boldsymbol{N}_\text{pointwise}$ refers to noise that is independent between individual pixels, originating from the detector's signal processing system \cite{calib1,calib2}, such as dark current and readout noise. Related studies on the calibration of noise in natural images \cite{calib1,calib2} have confirmed that Tukey's lambda distribution with a shape parameter provides a more accurate estimation of pointwise noise, \emph{i.e.},
 \begin{equation}
     \boldsymbol{N}_\text{pointwise} \sim \text{TL}(0,\sigma_\text{pointwise}, \lambda_\text{pointwise}),
 \end{equation}
 where $\sigma_\text{pointwise}$ is the standard deviation and $\lambda_\text{pointwise}$ is the shape parameter. Pointwise noise can be obtained by calculating the difference between the original image and the row-wise blurred image, as shown in Figure~\ref{fig:calib}\textcolor{red}{(f)}. Figure~\ref{fig:calib}\textcolor{red}{(g)} illustrates the variation of two parameters with pixel values. The standard deviation is proportional to the pixel value, while the shape parameter is a small positive constant number. Therefore, the slope and intercept of the standard deviation $\sigma_\text{pointwise}$, along with the constant shape parameter $\lambda_\text{pointwise}$, can be calculated for data simulation.

 \begin{figure}
\centering
\includegraphics[width=.9\linewidth]{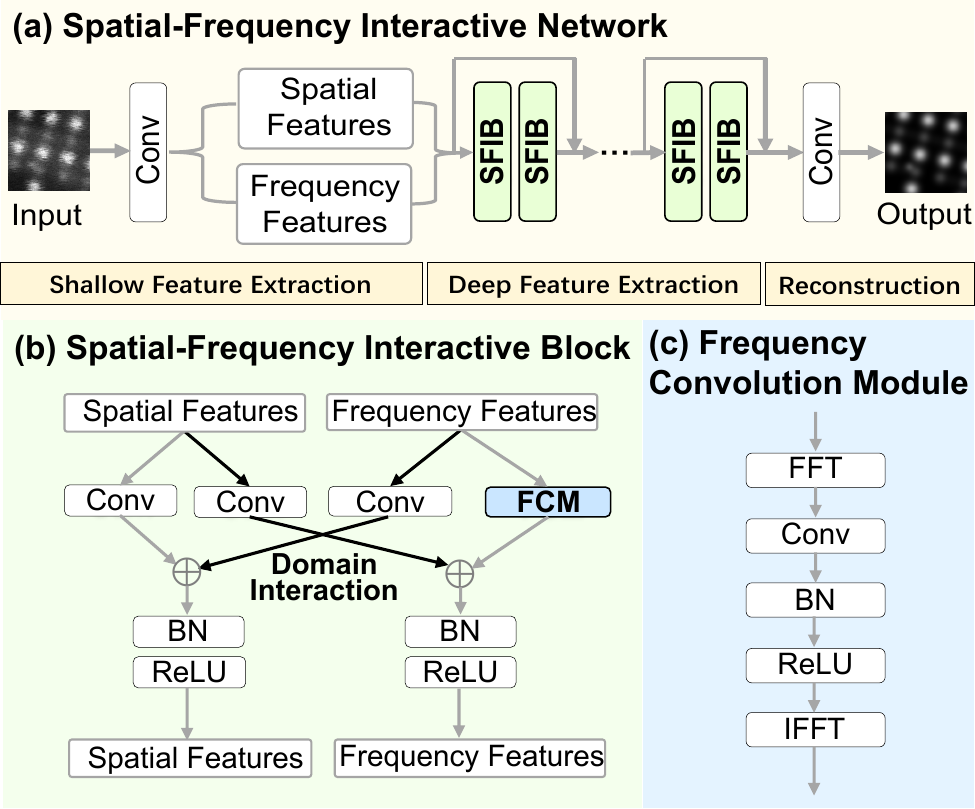}
\vspace{-2mm}
\caption{The framework of our Spatial-Frequency Interactive Network (SFIN) and the Spatial-Frequency Interactive Block (SFIB) and Frequency Convolution Module (FCM) used in it. Each spatial-frequency interactive block processes spatial and frequency domain features interactively. BN means batch normalization and IFFT means inverse fast Fourier transform.}
\vspace{-3mm}
\label{fig:frame}
\end{figure}

\subsection{Data Synthesis for STEM Image Enhancement}

Firstly, we calibrate the parameters of 40 real STEM image data (20 for HAADF mode and 20 for BF mode) with a size of 256$\times$256 according to our calibration method. The parameters include the minimum and maximum atomic brightness ($b_\text{atom}$), minimum and maximum background brightness ($b_\text{background}$), the slope and intercept of the scan noise standard deviation ($\sigma_\text{scan}$), the slope and intercept of the pointwise noise standard deviation ($\sigma_\text{pointwise}$), and the shape parameter of the pointwise noise ($\lambda_\text{pointwise}$). Each parameter varies slightly across different images, with their means and variances recorded. During data synthesis, each parameter is generated based on a normal distribution, followed by the generation of clean images and noise images using these parameters.

Then, we synthesize STEM image data in HAADF mode and BF mode using the calibrated parameters. Each mode includes 1000 groups of training data and 100 groups of testing data. Each group includes noisy input images and ground truth for the STEM image enhancement task. Each atomic column in the clean image is generated using a two-dimensional normal distribution. Their brightness and size relative to each other is rendered by the abTem library \cite{abTEM} according to the atomic structure. The overall brightness of the clean image is then scaled using the atom brightness parameter $b_\text{atom}$. Background noise is generated using two-dimensional Perlin noise \cite{Perlin}, with upsampling performed via bicubic interpolation. Scan noise is simulated using one-dimensional Perlin noise \cite{Perlin}, achieved by linearly combining two noise patterns with different standard deviations and changing with the amplitude of the image gradient. Pointwise noise is generated using Tukey's lambda distributions that are independent across pixels. Within the same image, the shape parameter $\lambda_\text{pointwise}$ remains constant while the standard deviation $\sigma_\text{pointwise}$ varies with pixel brightness.

In addition to the 13 material structures considered in existing datasets \cite{AtomSegNet,GAN}, we include other two common material structures, $\text{Sb}_2 \text{Se}_3$ and GaAs. Our dataset also contains three kinds of random structures, as shown in Figure~\ref{fig:random}. The first scenario simulates defects by deleting atoms randomly. The second adds atoms randomly to simulate embedding. The third totally randomizes atomic positions and sizes to simulate interfaces between different structures. 

 \begin{figure}
\centering
\includegraphics[width=1.\linewidth]{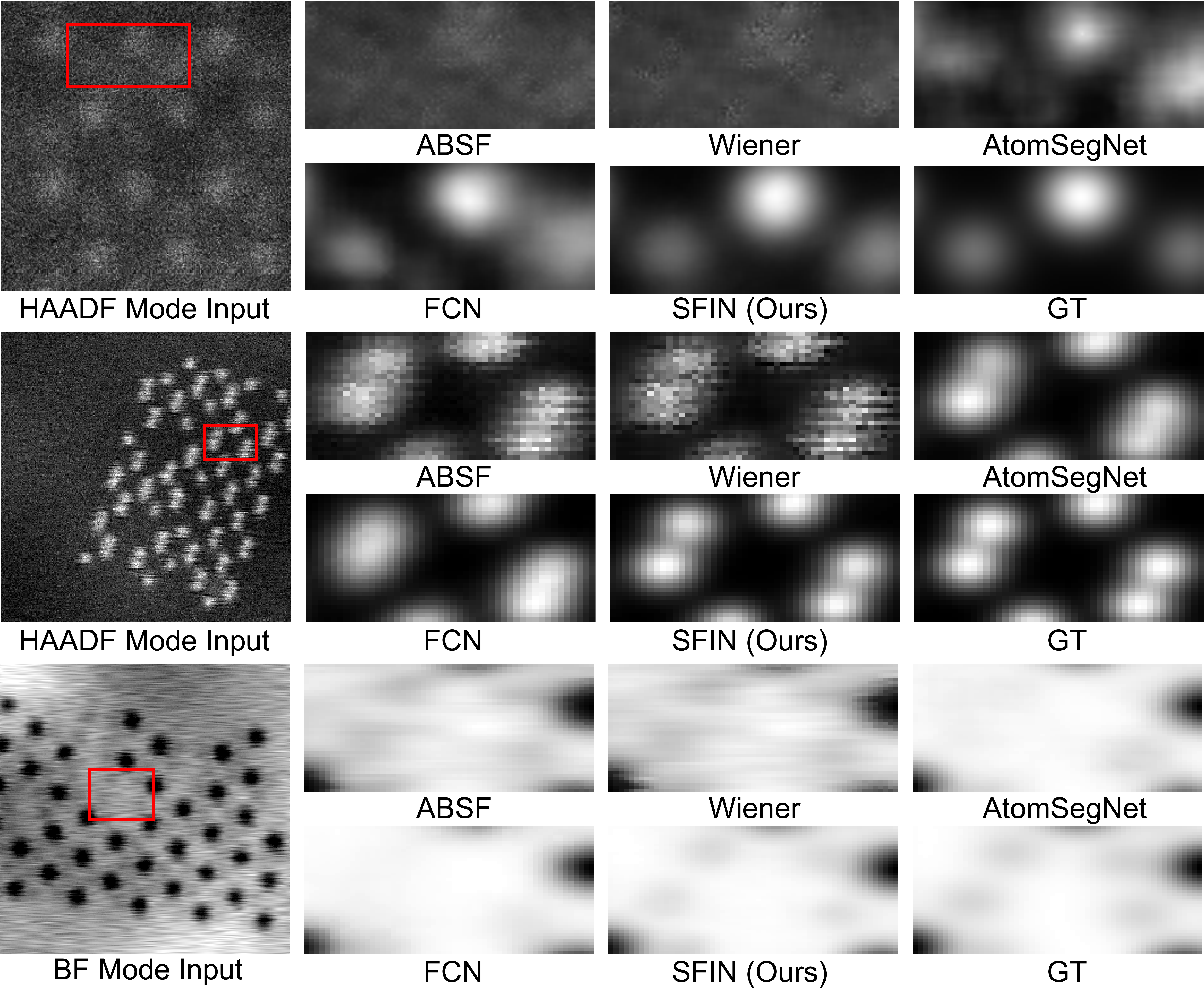}
\vspace{-8mm}
\caption{STEM image enhancement results of different methods. Comparison methods are ABSF \cite{Bilateral}, Wiener filter \cite{Wiener}, AtomSegNet \cite{AtomSegNet}, and FCN \cite{FCN}. In the first row, models are trained and tested on TEMImageNet \cite{AtomSegNet}. In the bottom two rows, models are trained and tested on our dataset.}
\vspace{-4mm}
\label{fig:enhance}
\end{figure}

\subsection{Spatial-Frequency Interactive Network}
\label{sec:network-design}
As analyzed in Section~\ref{sec:moti}, combining information from frequency and spatial domains considers both the regular arrangement and random variations of atoms, thereby enhancing the overall effect. As shown in Figure~\ref{fig:frame}\textcolor{red}{(a)}, our spatial-frequency interactive network processes input images in three steps, including shallow feature extraction, deep feature extraction, and reconstruction. 

In shallow feature extraction, a convolution is used to preliminarily extract feature maps. The extracted feature map is divided into two parts with the same number of channels, which are used to further extract spatial and frequency domain features, respectively.

In deep feature extraction, a series of spatial-frequency interactive blocks are used to gradually enhance the extracted shallow features. As shown in Figure~\ref{fig:frame}\textcolor{red}{(b)}, the spatial features and frequency features are enhanced through a convolution and a frequency convolution module, respectively. To ensure information exchange between the spatial and frequency domains, two additional convolution operations are introduced and the computed results are added to the results of the opposite domains. As shown in Figure~\ref{fig:frame}\textcolor{red}{(c)}, frequency domain processing of feature maps in a frequency convolution module includes a Fourier transform, a convolution, a batch normalization, a ReLU activation, and an inverse Fourier transform.

In reconstruction, convolution is used to synthesize the extracted spatial and frequency domain features into the output image of STEM image enhancement.

\section{Experiments}
\label{sec:exp}
 
In this section, we first provide the experimental settings and then present the results of different STEM image enhancement methods. We conduct ablation studies to verify the effectiveness of our dataset and frequency domain processing used in our network. Finally, we conduct extra experiments to validate the effectiveness of our method on atom detection and atom supper-resolution tasks. 

 \begin{figure*}
    \centering
    \includegraphics[width=.95\linewidth]{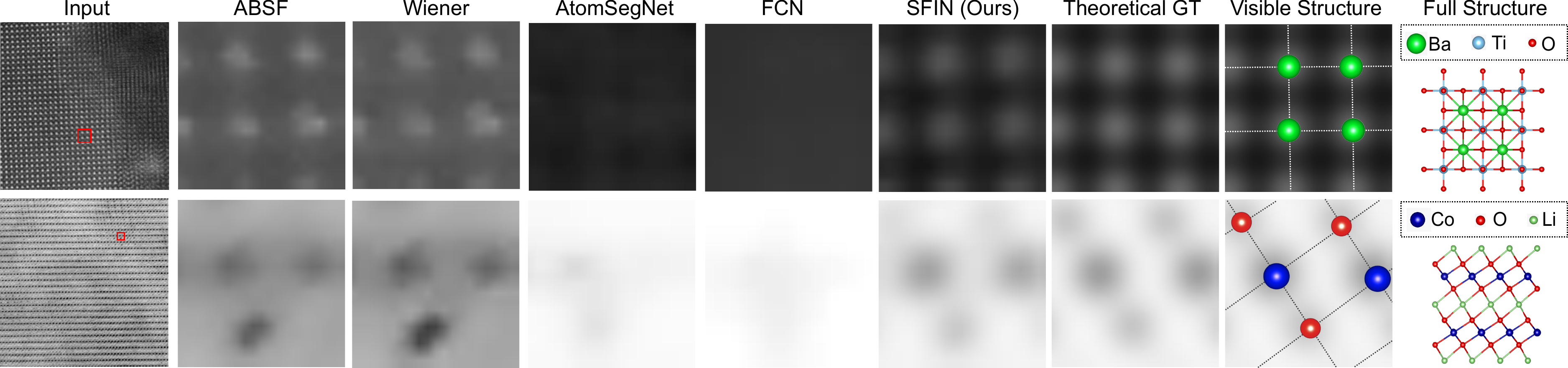}
    \vspace{-3.5mm}
    \caption{The enhancement results of real STEM images obtained by different methods trained on our dataset. Comparison methods are ABSF \cite{Bilateral}, Wiener filter \cite{Wiener}, AtomSegNet \cite{AtomSegNet}, and FCN \cite{FCN}. The theoretical GTs are noise-free STEM images rendered based on molecular structure.}
    \label{fig:real}
    \vspace{-5mm}
\end{figure*}

  \begin{figure}
    \centering
    \includegraphics[width=1\linewidth]{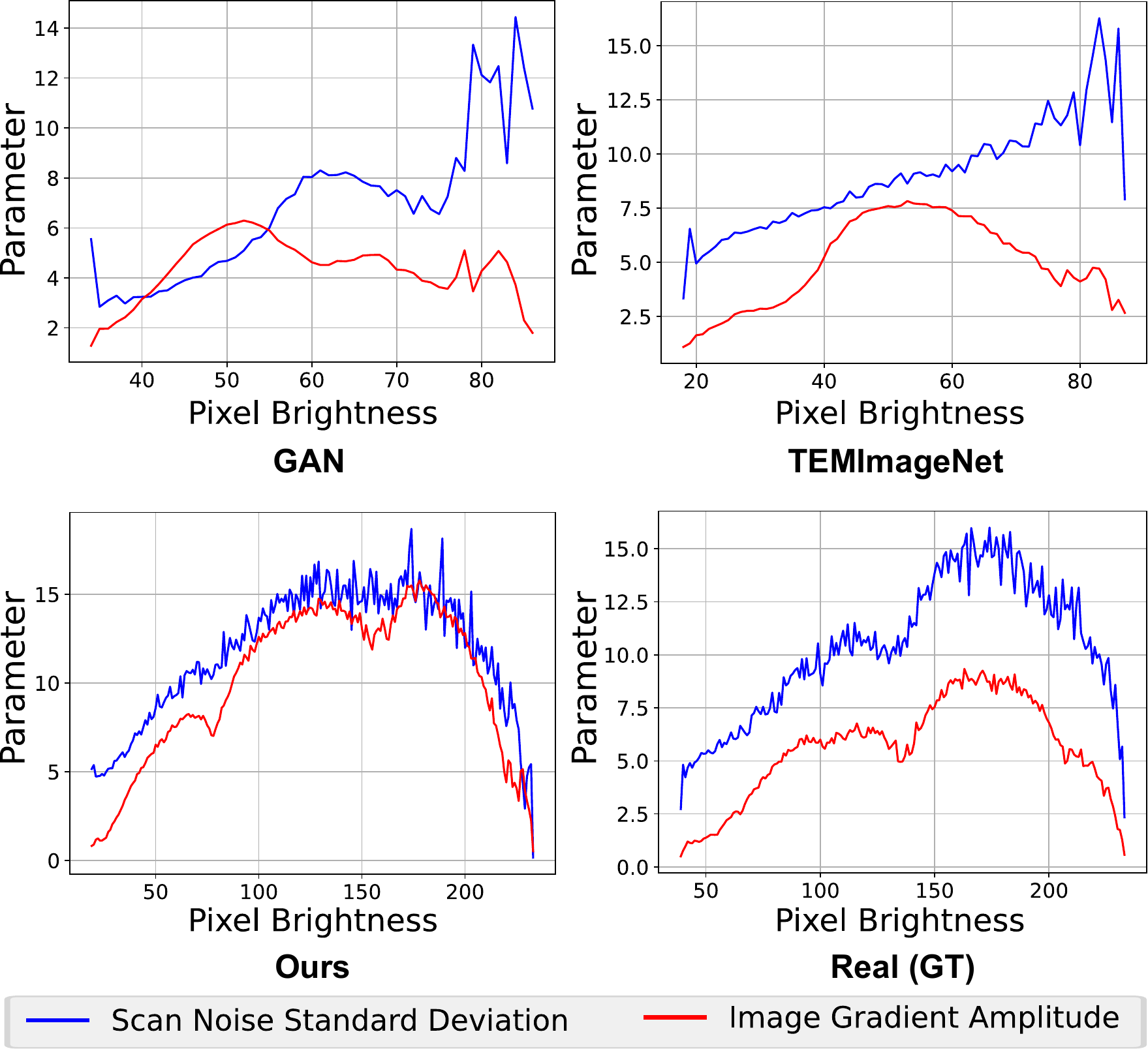}
    \vspace{-5mm}
    \caption{Trend comparison of scan noise standard deviation and image gradient amplitude in different datasets (HAADF mode). Comparison datasets are GAN \cite{GAN}, TEMImageNet \cite{AtomSegNet}, and real data (GT). All the gradient amplitude curves are divided by 5 for a better view. Only our dataset exhibits a similar trend with real data that scan noise standard deviation is proportional to image gradient amplitude.}
    \label{fig:curve}
    \vspace{-5mm}
\end{figure}

\subsection{Experimental Settings}
We first introduce the training method of our spatial-frequency interactive network. Then we provide the comparison datasets, comparison methods, and corresponding metrics used in the experiments.
 
 \vspace{5pt}\noindent\textbf{Training details of our network.} 
 Our spatial-frequency interactive network is trained with Adam optimizer \cite{Adam} ($\beta_1$ = 0.9 and $\beta_2$ = 0.999) and $L_1$ loss for 500 epochs on the dataset we synthesized. The codes are based on PyTorch \cite{Pytorch}. The initial learning rate is set as $2\times10^{-4}$ and reduced by half at the [250, 400, 450, 475]-th epochs. Mini-batch size is set to 8, which ensures that training can finish on a single NVIDIA 3090 GPU.

 \begin{table}[]\small
\centering
\caption{STEM image enhancement results of different methods trained and tested on the same dataset. The best and second best values are marked in \textbf{bold} and {\ul underlined}, respectively.}
\vspace{-2mm}
\resizebox{1\columnwidth}{!}{
\setlength{\tabcolsep}{1.mm}
\begin{NiceTabular}{c|cc|cc|cc}
\hline
                             & \multicolumn{2}{c|}{TEMImageNet \cite{AtomSegNet}} & \multicolumn{2}{c|}{Our Dataset}  & \multicolumn{2}{c}{Our Dataset}  \\
                             & \multicolumn{2}{c|}{(HAADF mode)}                  & \multicolumn{2}{c|}{(HAADF mode)} & \multicolumn{2}{c}{(BF mode)}    \\ \cline{2-7} 
\multirow{-3}{*}{Method}     & PSNR $\uparrow$           & SSIM $\uparrow$            & PSNR $\uparrow$   & SSIM $\uparrow$   & PSNR $\uparrow$  & SSIM $\uparrow$   \\ \hline
ABSF \cite{Bilateral}        & 15.98                   & 0.5204                   & 16.40           & 0.3658          & 10.32          & 0.7949          \\
Wiener Filter \cite{Wiener}  & 16.08                   & 0.4991                   & 16.43           & 0.3267          & 10.23          & 0.8081          \\
AtomSegNet \cite{AtomSegNet} & 22.34                   & 0.8008                   & 34.33           & 0.8994          & 30.26          & 0.9673          \\
FCN \cite{FCN}               & {\ul 26.18}             & {\ul 0.8842}             & {\ul 36.50}     & {\ul 0.9449}    & {\ul 32.22}    & {\ul 0.9847}    \\
\rowcolor[HTML]{F2F2F2} 
SFIN (Ours)                  & \textbf{31.76}          & \textbf{0.9543}          & \textbf{38.48}  & \textbf{0.9582} & \textbf{33.24} & \textbf{0.9885} \\ \hline
\end{NiceTabular}}\label{tab:enhance}
\vspace{-5mm}
 \end{table}

\vspace{5pt}\noindent\textbf{Comparison datasets and corresponding metrics.} 
Our dataset is compared with two existing synthetic STEM datasets and collected real datasets. The two existing datasets are GAN \cite{GAN} and TEMImageNet \cite{AtomSegNet}, both of which only contain HAADF mode images. The real dataset comes from several unprocessed real STEM images we collect, including HAADF mode and BF mode. All calibration parameters mentioned in section~\ref{sec:calib} are used to compare the similarity of distributions across different datasets. Kullback-Leibler divergence and $R^2$ metrics between synthetic datasets and real datasets are used to measure the realism of synthetic datasets.

\vspace{5pt}\noindent\textbf{Comparison methods and corresponding metrics.} 
Our spatial-frequency interactive network is compared with two traditional methods and two deep learning-based methods for STEM image enhancement. The two traditional methods are commonly used enhancement methods for STEM images, \emph{i.e.}, Wiener filtering \cite{Wiener}, and bilateral filtering \cite{Bilateral}. The two deep learning-based methods are FCN \cite{FCN} and AtomSegNet \cite{AtomSegNet}, which are state-of-the-art methods based on fully convolutional network and UNet \cite{UNet}, respectively. Peak Signal-to-Noise Ratio (PSNR) and Structural Similarity Index Measure (SSIM) \cite{SSIM} are used as the metrics.

 \begin{figure}
\centering
\includegraphics[width=.85\linewidth]{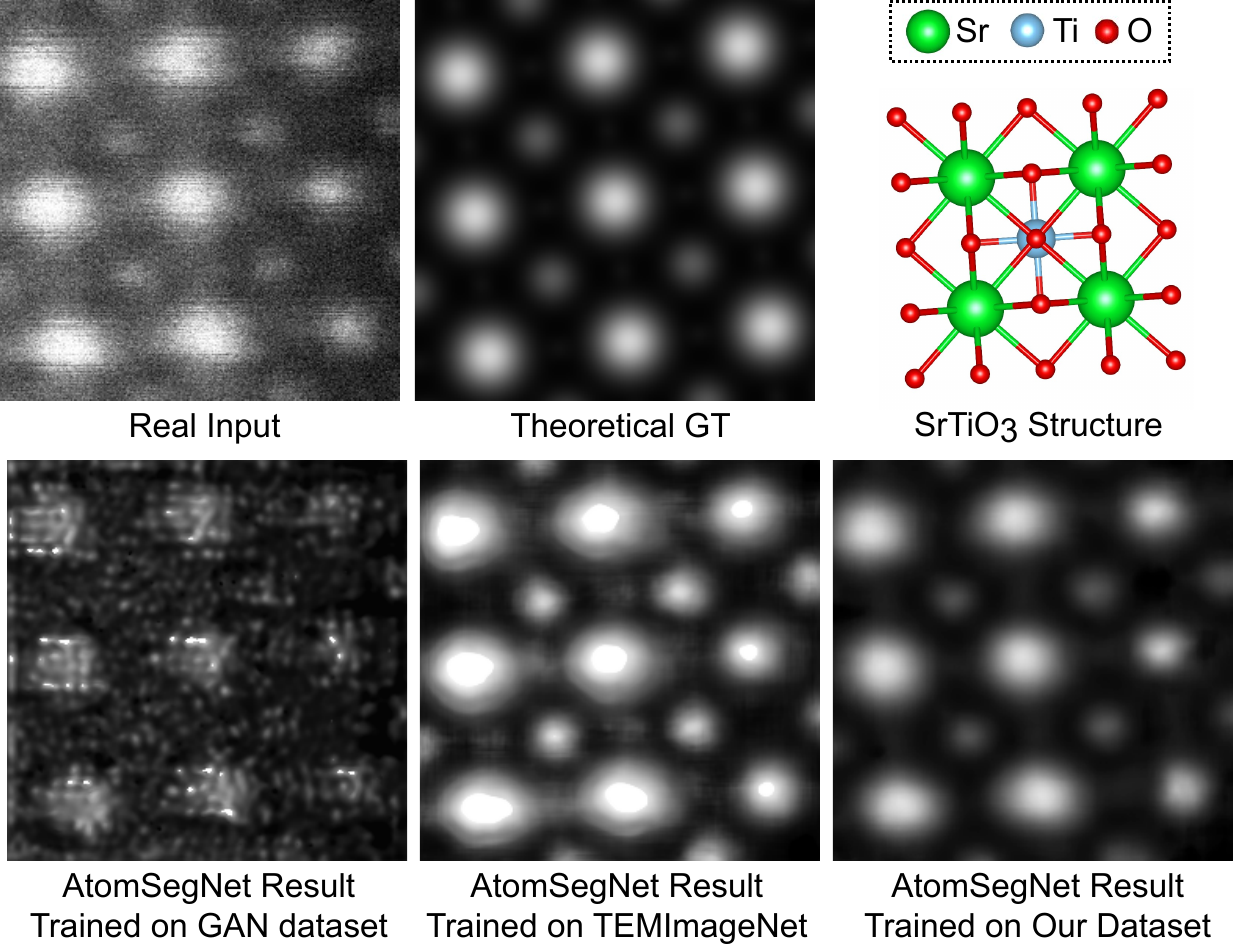}
\vspace{-3mm}
\caption{Enhancement results of real STEM image (HAADF mode). The results are obtained by AtomSegNet \cite{AtomSegNet} models trained on different datasets, including GAN \cite{GAN}, TEMImageNet \cite{AtomSegNet}, and ours. The theoretical GT is a noise-free STEM image rendered based on molecular structure. Results of GAN and TEMImageNet datasets show overexposure and unevenness, while our dataset yields a better result more similar to the theoretical GT.}
\vspace{-5mm}
\label{fig:real-data}
\end{figure}

\subsection{Main Result} We evaluate our Spatial-Frequency Interactive Network (SFIN) on both synthetic and real datasets. 

\begin{table*}[]\small
\centering
\caption{Calibration results of different datasets. The parameters that are closer to the parameters of the real (GT) dataset are better. Best and second best values are \textbf{bold} and {\ul underlined}, respectively. GAN \cite{GAN} and TEMImageNet \cite{AtomSegNet} datasets do not contain BF mode images.}
\vspace{-3mm}
\resizebox{.65\textwidth}{!}{
\setlength{\tabcolsep}{1.mm}
\begin{NiceTabular}{c|c|cc|cc|cc|ccc}
\hline
                        &                               & \multicolumn{2}{c|}{Atom Brightness}                                             & \multicolumn{2}{c|}{Background Noise}                                           & \multicolumn{2}{c|}{Scan Nosie}                                               & \multicolumn{3}{c}{Pointwise Noise}                                                                                    \\
\multirow{-2}{*}{Mode}  & \multirow{-2}{*}{Dataset}     & Min                                    & Max                                     & Min                                    & Max                                    & $k$                                   & $b$                                   & $k$                                   & $b$                                    & $\lambda$                             \\ \hline
                        & GAN \cite{GAN}                & 51.64                                  & 111.83                                  & 29.44                                  & {\ul 43.20}                            & -0.05                                 & 8.63                                  & {\ul 0.29}                            & -1.61                                  & 0.16                                  \\
                        & TEMImageNet \cite{AtomSegNet} & {\ul 62.82}                            & {\ul 247.18}                            & {\ul 0.00}                             & 36.66                                  & {\ul 0.13}                            & {\ul 4.31}                            & 0.33                                  & {\ul 8.14}                             & {\ul 0.34}                            \\
                        & \cellcolor[HTML]{F2F2F2}Ours  & \cellcolor[HTML]{F2F2F2}\textbf{63.12} & \cellcolor[HTML]{F2F2F2}\textbf{231.84} & \cellcolor[HTML]{F2F2F2}\textbf{10.05} & \cellcolor[HTML]{F2F2F2}\textbf{54.33} & \cellcolor[HTML]{F2F2F2}\textbf{0.11} & \cellcolor[HTML]{F2F2F2}\textbf{4.73} & \cellcolor[HTML]{F2F2F2}\textbf{0.08} & \cellcolor[HTML]{F2F2F2}\textbf{11.01} & \cellcolor[HTML]{F2F2F2}\textbf{0.21} \\ \cline{2-11} 
\multirow{-4}{*}{HAADF} & Real (GT)                     & 63.44                                  & 229.32                                  & 10.38                                  & 84.23                                  & 0.11                                  & 4.76                                  & 0.07                                  & 11.35                                  & 0.27                                  \\ \hline
                        & \cellcolor[HTML]{F2F2F2}Ours  & \cellcolor[HTML]{F2F2F2}208.55         & \cellcolor[HTML]{F2F2F2}57.62           & \cellcolor[HTML]{F2F2F2}116.88         & \cellcolor[HTML]{F2F2F2}243.87         & \cellcolor[HTML]{F2F2F2}0.02          & \cellcolor[HTML]{F2F2F2}8.73          & \cellcolor[HTML]{F2F2F2}0.01          & \cellcolor[HTML]{F2F2F2}16.18          & \cellcolor[HTML]{F2F2F2}0.39          \\ \cline{2-11} 
\multirow{-2}{*}{BF}    & Real (GT)                     & 227.85                                 & 51.60                                   & 121.24                                 & 242.45                                 & 0.03                                  & 7.85                                  & 0.01                                  & 17.03                                  & 0.28                                  \\ \hline
\end{NiceTabular}}\label{tab:calib}
\vspace{-2mm}
 \end{table*}

  \begin{table*}[]\small
\centering
\caption{The similarity between different datasets and real data. The best and second best values are marked in \textbf{bold} and {\ul underlined}, respectively. GAN \cite{GAN} and TEMImageNet \cite{AtomSegNet} datasets do not contain BF mode images.}
\vspace{-3mm}
\resizebox{0.95\textwidth}{!}{
\setlength{\tabcolsep}{1.mm}
\begin{NiceTabular}{c|c|cc|cc|cccc|cccccc}
\hline
                        &                               & \multicolumn{2}{c|}{}                                                         & \multicolumn{2}{c|}{Background}                                                & \multicolumn{4}{c|}{Scan Nosie}                                                                                                                                              & \multicolumn{6}{c}{Pointwise Noise}                                                                                                                                                                                                                                                 \\ \cline{7-16} 
                        &                               & \multicolumn{2}{c|}{\multirow{-2}{*}{Atom Brightness}}                        & \multicolumn{2}{c|}{Brightness}                                                & \multicolumn{2}{c|}{$k$}                                                                        & \multicolumn{2}{c|}{$b$}                                                   & \multicolumn{2}{c|}{$k$}                                                                           & \multicolumn{2}{c|}{$b$}                                                                          & \multicolumn{2}{c}{$\lambda$}                                              \\ \cline{3-16} 
\multirow{-3}{*}{Mode}  & \multirow{-3}{*}{Dataset}     & KLD $\downarrow$                        & $R^2$ $\uparrow$                        & KLD $\downarrow$                        & $R^2$ $\uparrow$                         & KLD $\downarrow$                     & \multicolumn{1}{c|}{$R^2$ $\uparrow$}                        & KLD $\downarrow$                     & $R^2$ $\uparrow$                        & KLD $\downarrow$                        & \multicolumn{1}{c|}{$R^2$ $\uparrow$}                        & KLD $\downarrow$                       & \multicolumn{1}{c|}{$R^2$ $\uparrow$}                        & KLD $\downarrow$                        & $R^2$ $\uparrow$                     \\ \hline
                        & GAN \cite{GAN}                & 15.05                                 & -6.56                                 & {\ul 8.34}                            & {\ul -1.74}                            & 21.58                              & \multicolumn{1}{c|}{-4.1}                                  & 21.83                              & -1.89                                 & 22.18                                 & \multicolumn{1}{c|}{-2.29}                                 & 21.22                                & \multicolumn{1}{c|}{-7.12}                                 & 16.07                                 & -2.62                              \\
                        & TEMImageNet \cite{AtomSegNet} & {\ul 4.66}                            & {\ul -1.11}                           & 10.12                                 & -7.26                                  & \textbf{0.56}                      & \multicolumn{1}{c|}{{\ul 0.26}}                            & \textbf{0.95}                      & {\ul 0.02}                            & {\ul 1.82}                            & \multicolumn{1}{c|}{{\ul -0.22}}                           & {\ul 9.41}                           & \multicolumn{1}{c|}{{\ul -1.06}}                           & {\ul 2.23}                            & \textbf{0.56}                      \\
\multirow{-3}{*}{HAADF} & \cellcolor[HTML]{F2F2F2}Ours  & \cellcolor[HTML]{F2F2F2}\textbf{1.26} & \cellcolor[HTML]{F2F2F2}\textbf{0.22} & \cellcolor[HTML]{F2F2F2}\textbf{2.25} & \cellcolor[HTML]{F2F2F2}\textbf{-0.16} & \cellcolor[HTML]{F2F2F2}{\ul 2.99} & \multicolumn{1}{c|}{\cellcolor[HTML]{F2F2F2}\textbf{0.65}} & \cellcolor[HTML]{F2F2F2}{\ul 1.55} & \cellcolor[HTML]{F2F2F2}\textbf{0.13} & \cellcolor[HTML]{F2F2F2}\textbf{0.06} & \multicolumn{1}{c|}{\cellcolor[HTML]{F2F2F2}\textbf{0.95}} & \cellcolor[HTML]{F2F2F2}\textbf{0.2} & \multicolumn{1}{c|}{\cellcolor[HTML]{F2F2F2}\textbf{0.65}} & \cellcolor[HTML]{F2F2F2}\textbf{2.17} & \cellcolor[HTML]{F2F2F2}{\ul 0.49} \\ \hline
BF                      & \cellcolor[HTML]{F2F2F2}Ours  & \cellcolor[HTML]{F2F2F2}3.41          & \cellcolor[HTML]{F2F2F2}-0.85         & \cellcolor[HTML]{F2F2F2}2.39          & \cellcolor[HTML]{F2F2F2}-1             & \cellcolor[HTML]{F2F2F2}1.43       & \multicolumn{1}{c|}{\cellcolor[HTML]{F2F2F2}0}             & \cellcolor[HTML]{F2F2F2}1.24       & \cellcolor[HTML]{F2F2F2}0.32          & \cellcolor[HTML]{F2F2F2}2.47          & \multicolumn{1}{c|}{\cellcolor[HTML]{F2F2F2}-0.05}         & \cellcolor[HTML]{F2F2F2}2.48         & \multicolumn{1}{c|}{\cellcolor[HTML]{F2F2F2}-0.15}         & \cellcolor[HTML]{F2F2F2}0.81          & \cellcolor[HTML]{F2F2F2}-0.07      \\ \hline
\end{NiceTabular}}\label{tab:r2} 
\vspace{0mm}
 \end{table*}

   \begin{table*}[]\small
\centering
\caption{Applicable situations of different STEM datasets.}
\vspace{-3mm}
\resizebox{0.85\textwidth}{!}{
\setlength{\tabcolsep}{1.mm}
\begin{NiceTabular}{c|cc|ccc|c|c}
\hline
Dataset                       & Random Scale        & Random Rotation     & Atomic Defects      & Ion Embeddings      & Interfaces          & Pattern Number & BF Mode             \\ \hline
GAN \cite{GAN}                & \XSolidBrush        & \XSolidBrush        & \Checkmark          & \XSolidBrush        & \XSolidBrush        & 1              & \XSolidBrush        \\
TEMImageNet \cite{AtomSegNet} & Discrete            & Discrete            & \XSolidBrush        & \XSolidBrush        & \XSolidBrush        & 13             & \XSolidBrush        \\
\rowcolor[HTML]{F2F2F2} 
Ours                          & \textbf{Continuous} & \textbf{Continuous} & \textbf{\Checkmark} & \textbf{\Checkmark} & \textbf{\Checkmark} & \textbf{15}    & \textbf{\Checkmark} \\ \hline
\end{NiceTabular}}\label{tab:situation}
 \vspace{-5mm}
 \end{table*}

\vspace{2pt}\noindent\textbf{Evaluation on synthetic data.} Our SFIN and four comparison methods (\emph{i.e.}, ABSF \cite{Bilateral}, Wiener \cite{Wiener}, AtomSegNet \cite{AtomSegNet}, and FCN \cite{FCN}) are tested on the TEMImageNet synthetic dataset \cite{AtomSegNet} and our synthetic dataset, respectively. As shown in Table~\ref{tab:enhance}, our method demonstrates superior enhancement performance on both the existing TEMImageNet dataset \cite{AtomSegNet} and our synthetic dataset. Our PSNR metrics increase by 1-5 dB compared to the current best method, FCN \cite{FCN}. Figure~\ref{fig:enhance} presents a visual comparison of three groups of enhancement results. It can be observed that traditional methods like ABSF \cite{Bilateral} and Wiener \cite{Wiener} cannot completely remove noise, particularly high-intensity scan noise, which is mistakenly preserved as image texture. In the first row of Figure~\ref{fig:enhance}, AtomSegNet \cite{AtomSegNet} incorrectly predicts the left atom's position, while both AtomSegNet and FCN incorrectly predict the size and brightness of the right atom. Our method accurately predicts the positions, sizes, and intensities of all atoms. In the second row, AtomSegNet and FCN erroneously merge two closely spaced atoms, whereas our method correctly distinguishes them. In the third row, AtomSegNet and FCN remove some lighter-colored atoms as noise, while they are correctly preserved in our enhancement result. In summary, our method effectively removes the noise while accurately restoring each atom.

\vspace{5pt}\noindent\textbf{Evaluation on real data.}
To verify the effectiveness of various STEM image enhancement methods in practical applications, we collect a set of real image data with known material structures for testing, which includes both HAADF and BF mode images. Theoretically rendered ground truth images are used as references for comparison. Since previous simulated datasets (\emph{i.e.}, TEMImageNet \cite{AtomSegNet} and GAN \cite{GAN}) do not include BF mode images, all deep learning methods are trained on our BF image dataset to enhance BF mode images. Figure~\ref{fig:real} shows the enhancement results of different methods. The results from the two traditional methods (\emph{i.e.}, ABSF \cite{Bilateral} and Wiener \cite{Wiener}) still contain noise and background color. Compared to other deep learning methods (\emph{i.e.}, AtomSegNet \cite{AtomSegNet} and FCN \cite{FCN}), our SFIN more clearly displays the atoms.

\subsection{Effectiveness of Our Dataset}

To verify the realism of our synthesized data, all the calibration parameters mentioned in section~\ref{sec:calib} are used to compare the distributions across different datasets. As shown in Table~\ref{tab:calib}, Table~\ref{tab:r2}, and Figure~\ref{fig:curve}, our dataset exhibits a parameter distribution that is closer to real data compared to other synthetic datasets, \emph{i.e.}, GAN \cite{GAN} and TEMImageNet \cite{AtomSegNet}.

We also train the same network (AtomSegNet \cite{AtomSegNet}) using different synthetic datasets to validate the actual effectiveness of the synthetic datasets. As shown in Figure~\ref{fig:real-data}, there is still significant noise in the result trained with the GAN dataset \cite{GAN}. In the result trained with the TEMImageNet dataset \cite{AtomSegNet}, each atom exhibits an irregular shape and there is overexposure at the center. The result trained with our dataset is closer to the ideal theoretical ground truth image, where each atom is uniformly spherical and noise-free. The proposed noise calibration method and data synthesis method make the simulated data closer to real data, thereby achieving better enhancement effects.

 Table~\ref{tab:situation} compares the applicable situations of different datasets. Our dataset considers more situations, including more continuous image deformations, more atomic arrangements, and more shooting modes.

   \begin{table}[]\small
\centering
\caption{STEM Image enhancement results (HAADF mode) of our network with and without frequency domain process and domain interaction. Models are trained and tested on our dataset. Best and second best values are \textbf{bold} and {\ul underlined}, respectively.}
\vspace{-3mm}
\resizebox{1.\columnwidth}{!}{
\setlength{\tabcolsep}{1.mm}
\begin{NiceTabular}{ccc|cc}
\hline
Spatial Domain & Frequency Domain & Domain Interaction & PSNR $\uparrow$  & SSIM $\uparrow$   \\ \hline
\Checkmark     & \XSolidBrush    & \XSolidBrush       & 34.61          & 0.9050          \\
\Checkmark     & \Checkmark      & \XSolidBrush       & {\ul 38.38}    & {\ul 0.9514}    \\
\rowcolor[HTML]{F2F2F2} 
\Checkmark     & \Checkmark      & \Checkmark         & \textbf{38.48} & \textbf{0.9582} \\ \hline
\end{NiceTabular}}\label{tab:ablation}
\vspace{-3mm}
 \end{table}

\subsection{Ablation on Frequency Domain Processing}
To validate the effectiveness of frequency domain processing, we remove it from our proposed network for comparison. The first variant processes only in the spatial domain, similar to existing STEM image enhancement methods like AtomSegNet \cite{AtomSegNet} and FCN \cite{FCN}. In this variant, the frequency domain convolutions in each module are replaced with standard spatial domain convolutions. The second variant retains frequency domain operations but removes the two convolutional operations for information exchange between the domains (the dark arrows in Figure~\ref{fig:frame}\textcolor{red}{(b)}). 

Results presented in Table~\ref{tab:ablation} show a significant performance drop when only processing spatial domain information, indicating that effectively leveraging frequency domain information can enhance STEM image quality. Furthermore, information interaction between spatial and frequency domains yields better results.

\subsection{Extra Evaluation on other tasks}
\vspace{-1mm}
We further validate the generality of the proposed method on the atom detection and super-resolution tasks.

  \begin{table}[]\small
\centering
\caption{Atom detection results of different methods trained and tested on our dataset. The best and second best values are marked in \textbf{bold} and {\ul underlined}, respectively.}
\vspace{-3mm}
\resizebox{.8\columnwidth}{!}{
\setlength{\tabcolsep}{1.mm}
\begin{NiceTabular}{c|cc|cc}
\hline
                             & \multicolumn{2}{c|}{HAADF Mode}                     & \multicolumn{2}{c}{BF Mode}      \\
\multirow{-2}{*}{Method}     & \multicolumn{1}{l}{PSNR $\uparrow$} & SSIM $\uparrow$   & PSNR $\uparrow$  & SSIM $\uparrow$   \\ \hline
AtomSegNet \cite{AtomSegNet} & 26.14                             & 0.8293          & 23.76          & 0.8065          \\
FCN \cite{FCN}               & {\ul 26.82}                       & {\ul 0.9367}    & {\ul 24.58}    & {\ul 0.9148}    \\
\rowcolor[HTML]{F2F2F2} 
SFIN (Ours)                  & \textbf{28.67}                    & \textbf{0.9642} & \textbf{25.77} & \textbf{0.9430} \\ \hline
\end{NiceTabular}}\label{tab:detect}
 \vspace{-2mm}
 \end{table}

 \begin{table}[]\small
\centering
\caption{Atom supper-resolution results of different methods trained and tested on TEMImageNet dataset \cite{AtomSegNet}. The best and second best values are marked in \textbf{bold} and {\ul underlined}, respectively.}
\vspace{-3mm}
\resizebox{.8\columnwidth}{!}{
\setlength{\tabcolsep}{1.mm}
\begin{NiceTabular}{c|cc
>{\columncolor[HTML]{F2F2F2}}c }
\hline
Metric        & AtomSegNet \cite{AtomSegNet} & FCN \cite{FCN} & SFIN (Ours)     \\ \hline
PSNR $\uparrow$ & 29.73                        & {\ul 32.11}    & \textbf{41.74}  \\
SSIM $\uparrow$ & 0.8398                       & {\ul 0.8943}   & \textbf{0.9933} \\ \hline
\end{NiceTabular}}\label{tab:sr}
\vspace{-5mm}
 \end{table}

\begin{figure}
\centering
\includegraphics[width=.9\linewidth]{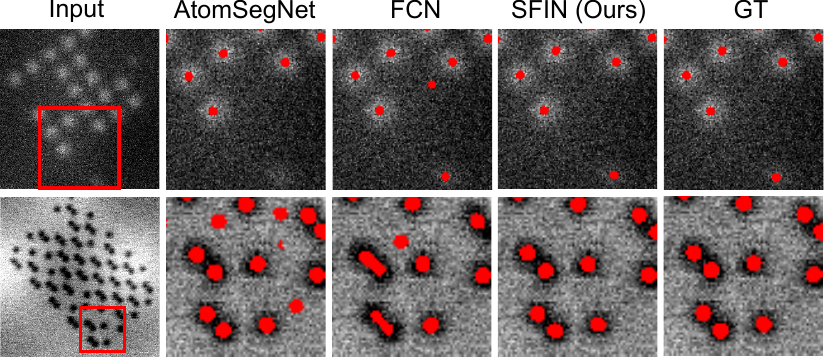}
\vspace{-3.5mm}
\caption{Atom detection results of different methods. Comparison methods are AtomSegNet \cite{AtomSegNet} and FCN \cite{FCN}. Models are trained and tested on our dataset.}
\vspace{-4mm}
\label{fig:detect}
\end{figure}

\vspace{2pt}\noindent\textbf{Atom detection.} Atom detection aims to identify each atom in STEM images and mark its central position with a dot. We synthesize the atom detection dataset like the synthetic STEM image enhancement dataset and then train and test different methods on the simulated dataset. As shown in Table~\ref{tab:detect}, our method outperforms AtomSegNet \cite{AtomSegNet} and FCN \cite{FCN}. To visually demonstrate the detection results of different methods, predicted labels are marked in red and overlaid on the input images. Results for one HAADF mode example are shown in the first row of Figure~\ref{fig:detect}. Our method exhibits fewer false detections and missed detections. Results for one BF mode example are shown in the second row of Figure~\ref{fig:detect}, where AtomSegNet mistakenly identifies some noise as atoms, and FCN incorrectly merges two nearby atoms into one. Our method exhibits fewer false detections and can distinguish closely spaced atoms more effectively.

\begin{figure}
\centering
\includegraphics[width=.9\linewidth]{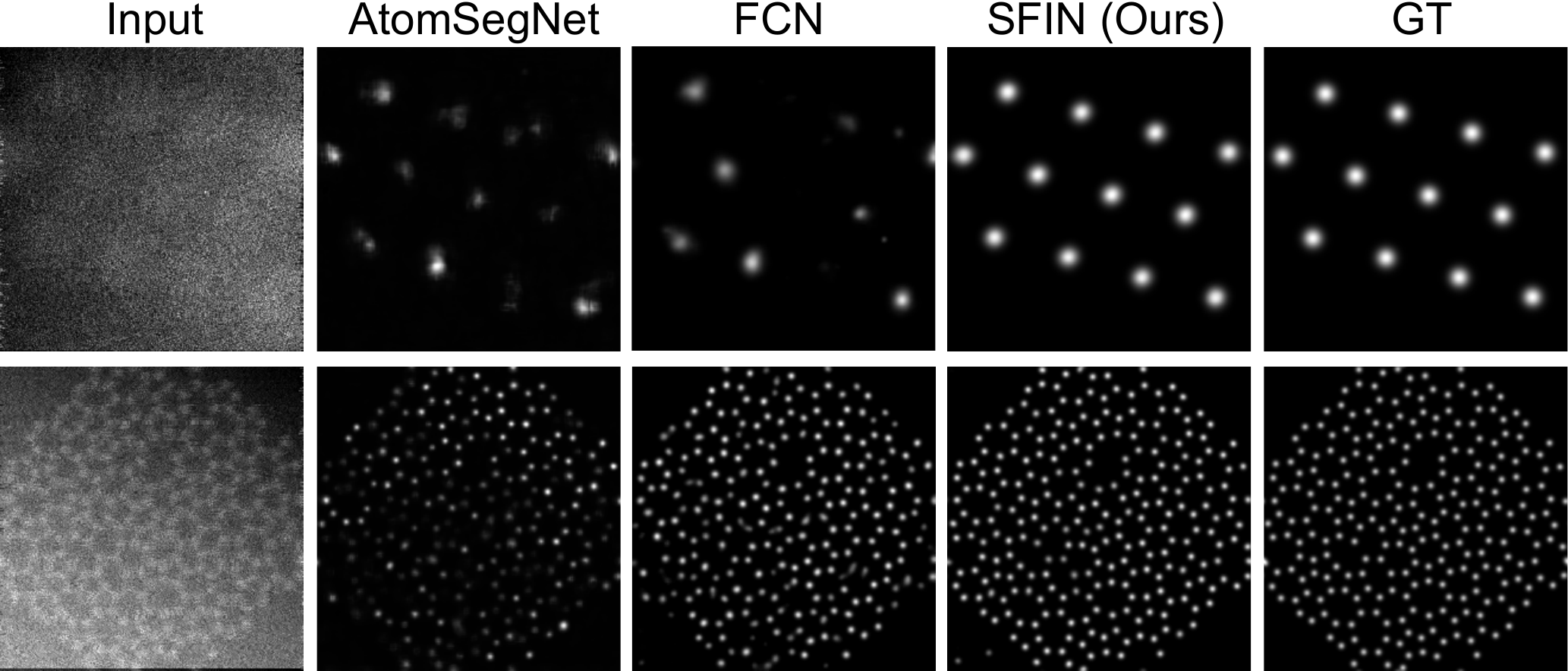}
\vspace{-3.5mm}
\caption{Atom super-resolution results of different methods. Comparison methods are AtomSegNet \cite{AtomSegNet} and FCN \cite{FCN}. Models are trained and tested on the TEMImageNet dataset \cite{AtomSegNet}.}
\vspace{-5mm}
\label{fig:sr}
\end{figure}

\vspace{2pt}\noindent\textbf{Atom super-resolution.} Atom super-resolution aims to enhance the clarity of original STEM images and reduce the atomic radius to facilitate distinguishing between different atoms and measuring interatomic distances. We train and test different methods on the atom super-resolution simulation data provided by the TEMImageNet dataset \cite{AtomSegNet}. As shown in Table~\ref{tab:sr}, our method outperforms AtomSegNet \cite{AtomSegNet} and FCN \cite{FCN}. The visualization results are presented in Figure~\ref{fig:sr}. When the input images are significantly blurred, our method successfully identifies each atom, whereas the results of other methods are incomplete.

\vspace{-1mm}
\section{Conclusion}
\vspace{-1mm}
We propose a noise calibration method for STEM images to synthesize a more realistic dataset and integrate frequency domain processing into networks for better performance. Additionally, we include various types of randomized samples in the dataset to simulate special atomic arrangements and provide both HAADF and BF mode images, which make the dataset more general. Experiments demonstrate that our method achieves superior quantitative and visual results. Ablation studies confirm the effectiveness of the proposed dataset and frequency domain processing. The proposed method also delivers better performance in other STEM image processing tasks such as atom detection and atom super-resolution. In the future, we will expand our dataset and experiments to encompass more tasks.

\vspace{-1mm}
\section*{Acknowledgements}
\vspace{-1mm}
This work is supported by the National Natural Science Foundation of China (62331006, 62171038, and 62088101), and the Fundamental Research Funds for the Central Universities.

{
    \small
    \bibliographystyle{ieeenat_fullname}
    \bibliography{main}

\begin{thebibliography}{30}
\providecommand{\natexlab}[1]{#1}
\providecommand{\url}[1]{\texttt{#1}}
\expandafter\ifx\csname urlstyle\endcsname\relax
  \providecommand{\doi}[1]{doi: #1}\else
  \providecommand{\doi}{doi: \begingroup \urlstyle{rm}\Url}\fi

\bibitem[Conrad and Narayan(2021)]{CEM500k}
Ryan Conrad and Kedar Narayan.
\newblock Cem500k, a large-scale heterogeneous unlabeled cellular electron microscopy image dataset for deep learning.
\newblock \emph{Elife}, 10:\penalty0 e65894, 2021.

\bibitem[Fu and Li(2025)]{FCDFusion}
Ying Fu and Hesong Li.
\newblock Fcdfusion: A fast, low color deviation method for fusing visible and infrared image pairs.
\newblock \emph{Computational Visual Media}, 11\penalty0 (1):\penalty0 195--211, 2025.

\bibitem[Fultz and Howe(2013)]{bf3}
Brent Fultz and James Howe.
\newblock Diffraction contrast in tem images.
\newblock \emph{Transmission Electron Microscopy and Diffractometry of Materials}, pages 349--427, 2013.

\bibitem[Kazimi et~al.(2024)Kazimi, Ruzaeva, and Sandfeld]{GAN2}
Bashir Kazimi, Karina Ruzaeva, and Stefan Sandfeld.
\newblock Self-supervised learning with generative adversarial networks for electron microscopy.
\newblock In \emph{Proceedings of the IEEE Conference on Computer Vision and Pattern Recognition Workshops}, pages 71--81, 2024.

\bibitem[Khan et~al.(2023)Khan, Lee, Huang, and Clark]{GAN}
Abid Khan, Chia-Hao Lee, Pinshane~Y Huang, and Bryan~K Clark.
\newblock Leveraging generative adversarial networks to create realistic scanning transmission electron microscopy images.
\newblock \emph{npj Computational Materials}, 9\penalty0 (1):\penalty0 85, 2023.

\bibitem[Kingma and Ba(2014)]{Adam}
Diederik~P Kingma and Jimmy Ba.
\newblock Adam: A method for stochastic optimization.
\newblock \emph{arXiv preprint arXiv:1412.6980}, 2014.

\bibitem[Lee et~al.(2020)Lee, Khan, Luo, Santos, Shi, Janicek, Kang, Zhu, Sobh, Schleife, et~al.]{FCN}
Chia-Hao Lee, Abid Khan, Di Luo, Tatiane~P Santos, Chuqiao Shi, Blanka~E Janicek, Sangmin Kang, Wenjuan Zhu, Nahil~A Sobh, Andr{\'e} Schleife, et~al.
\newblock Deep learning enabled strain mapping of single-atom defects in two-dimensional transition metal dichalcogenides with sub-picometer precision.
\newblock \emph{Nano Letters}, 20\penalty0 (5):\penalty0 3369--3377, 2020.

\bibitem[Li et~al.(2023)Li, Dai, Zhao, Ma, Cao, and Zhang]{CJE1}
Hongliang Li, Feng Dai, Qiang Zhao, Yike Ma, Juan Cao, and Yongdong Zhang.
\newblock Non-uniform compressive sensing imaging based on image saliency.
\newblock \emph{Chinese Journal of Electronics}, 32\penalty0 (1):\penalty0 159--165, 2023.

\bibitem[Lin et~al.(2021)Lin, Zhang, Wang, Yang, and Xin]{AtomSegNet}
Ruoqian Lin, Rui Zhang, Chunyang Wang, Xiao-Qing Yang, and Huolin~L Xin.
\newblock Temimagenet training library and atomsegnet deep-learning models for high-precision atom segmentation, localization, denoising, and deblurring of atomic-resolution images.
\newblock \emph{Scientific Reports}, 11\penalty0 (1):\penalty0 5386, 2021.

\bibitem[Madsen and Susi(2021)]{abTEM}
Jacob Madsen and Toma Susi.
\newblock The abtem code: transmission electron microscopy from first principles.
\newblock \emph{Open Research Europe}, 1:\penalty0 24, 2021.

\bibitem[Paszke et~al.(2019)Paszke, Gross, Massa, Lerer, and Chintala]{Pytorch}
Adam Paszke, Sam Gross, Francisco Massa, Adam Lerer, and Soumith Chintala.
\newblock Pytorch: An imperative style, high-performance deep learning library.
\newblock 2019.

\bibitem[Pennycook and Jesson(1991)]{bf2}
S.J. Pennycook and D.E. Jesson.
\newblock High-resolution z-contrast imaging of crystals.
\newblock \emph{Ultramicroscopy}, 37\penalty0 (1):\penalty0 14--38, 1991.

\bibitem[Pennycook et~al.(2003)Pennycook, Jesson, Chisholm, Browning, McGibbon, and McGibbon]{bf1}
SJ Pennycook, DE Jesson, MF Chisholm, ND Browning, AJ McGibbon, and MM McGibbon.
\newblock {Z-Contrast Imaging in the Scanning Transmission Electron Microscope}.
\newblock \emph{Microscopy and Microanalysis}, 1\penalty0 (6):\penalty0 231--251, 2003.

\bibitem[Perlin(1985)]{Perlin}
Ken Perlin.
\newblock An image synthesizer.
\newblock In \emph{Proceedings of Conference on Computer Graphics and Interactive Techniques}, pages 287--296, 1985.

\bibitem[Ronneberger et~al.(2015)Ronneberger, Fischer, and Brox]{UNet}
Olaf Ronneberger, Philipp Fischer, and Thomas Brox.
\newblock U-net: Convolutional networks for biomedical image segmentation.
\newblock In \emph{Proceedings of Medical image computing and computer-assisted intervention}, pages 234--241, 2015.

\bibitem[Sánchez-Santolino et~al.(2024)Sánchez-Santolino, Rouco, Puebla, Aramberri, Zamora, Cabero, Cuellar, Munuera, Mompean, Garcia-Hernandez, Castellanos-Gomez, Íñiguez, Leon, and Santamaria]{semiconductor}
G. Sánchez-Santolino, V. Rouco, S. Puebla, H. Aramberri, V. Zamora, M. Cabero, F.~A. Cuellar, C. Munuera, F. Mompean, M. Garcia-Hernandez, A. Castellanos-Gomez, J. Íñiguez, C. Leon, and J. Santamaria.
\newblock A 2d ferroelectric vortex pattern in twisted batio3 freestanding layers.
\newblock \emph{Nature}, 626\penalty0 (7999):\penalty0 529--534, 2024.

\bibitem[Tian et~al.(2023)Tian, Fu, and Zhang]{CJE3}
Ye Tian, Ying Fu, and Jun Zhang.
\newblock Transformer-based under-sampled single-pixel imaging.
\newblock \emph{Chinese Journal of Electronics}, 32\penalty0 (5):\penalty0 1151--1159, 2023.

\bibitem[Tomasi and Manduchi(1998)]{Bilateral}
Carlo Tomasi and Roberto Manduchi.
\newblock Bilateral filtering for gray and color images.
\newblock In \emph{Proceedings of the IEEE International Conference on Computer Vision}, pages 839--846, 1998.

\bibitem[Wang et~al.(2023)Wang, Wang, Zhang, Lei, Kisslinger, and Xin]{Battery2}
Chunyang Wang, Xuelong Wang, Rui Zhang, Tianjiao Lei, Kim Kisslinger, and Huolin~L. Xin.
\newblock Resolving complex intralayer transition motifs in high-ni-content layered cathode materials for lithium-ion batteries.
\newblock \emph{Nature Materials}, 22\penalty0 (2), 2023.

\bibitem[Wang et~al.(2024)Wang, Jing, Zhu, and Xin]{Battery1}
Chunyang Wang, Yaqi Jing, Dong Zhu, and Huolin~L. Xin.
\newblock Atomic origin of chemomechanical failure of layered cathodes in all-solid-state batteries.
\newblock \emph{Journal of the American Chemical Society}, 146\penalty0 (26):\penalty0 17712--17718, 2024.

\bibitem[Wang et~al.()Wang, Feng, Tang, Zhu, Cao, Zou, Geng, and Ma]{semiconductor2}
Yu-Jia Wang, Yan-Peng Feng, Yun-Long Tang, Yin-Lian Zhu, Yi Cao, Min-Jie Zou, Wan-Rong Geng, and Xiu-Liang Ma.
\newblock Polar bloch points in strained ferroelectric films.
\newblock \emph{Nature Communications}, 15\penalty0 (1):\penalty0 3949.

\bibitem[Wang et~al.(2004)Wang, Bovik, Sheikh, and Simoncelli]{SSIM}
Zhou Wang, Alan~C Bovik, Hamid~R Sheikh, and Eero~P Simoncelli.
\newblock Image quality assessment: from error visibility to structural similarity.
\newblock \emph{IEEE Transactions on Image Processing}, 13\penalty0 (4):\penalty0 600--612, 2004.

\bibitem[Wei et~al.(2020)Wei, Fu, Yang, and Huang]{calib1}
Kaixuan Wei, Ying Fu, Jiaolong Yang, and Hua Huang.
\newblock A physics-based noise formation model for extreme low-light raw denoising.
\newblock In \emph{Proceedings of the IEEE Conference on Computer Vision and Pattern Recognition}, pages 2758--2767, 2020.

\bibitem[Wei et~al.(2021)Wei, Fu, Zheng, and Yang]{calib2}
Kaixuan Wei, Ying Fu, Yinqiang Zheng, and Jiaolong Yang.
\newblock Physics-based noise modeling for extreme low-light photography.
\newblock \emph{IEEE Transactions on Pattern Analysis and Machine Intelligence}, 44\penalty0 (11):\penalty0 8520--8537, 2021.

\bibitem[Wiener(1949)]{Wiener}
Norbert Wiener.
\newblock \emph{Extrapolation, interpolation, and smoothing of stationary time series: with engineering applications}.
\newblock The MIT press, 1949.

\bibitem[Yang et~al.(2023)Yang, Zhang, Fu, Wang, Teng, Shao, Wu, Chang, Ding, Wang, and Han]{functional}
Chengpeng Yang, Bozhao Zhang, Libo Fu, Zhanxin Wang, Jiao Teng, Ruiwen Shao, Ziqi Wu, Xiaoxue Chang, Jun Ding, Lihua Wang, and Xiaodong Han.
\newblock Chemical inhomogeneity–induced profuse nanotwinning and phase transformation in aucu nanowires.
\newblock \emph{Nature Communications}, 14\penalty0 (1):\penalty0 5705, 2023.

\bibitem[Yanshan et~al.(2023)Yanshan, Shifu, Wenhan, Li, and Weixin]{CJE2}
LI Yanshan, CHEN Shifu, LUO Wenhan, ZHOU Li, and XIE Weixin.
\newblock Hyperspectral image super-resolution based on spatial-spectral feature extraction network.
\newblock \emph{Chinese Journal of Electronics}, 32\penalty0 (3):\penalty0 415--428, 2023.

\bibitem[Zhang et~al.(2023)Zhang, Wang, Zou, Lin, Ma, Li, Hwang, Xu, Sun, Trask, and Xin]{Battery3}
Rui Zhang, Chunyang Wang, Peichao Zou, Ruoqian Lin, Lu Ma, Tianyi Li, In-hui Hwang, Wenqian Xu, Chengjun Sun, Steve Trask, and Huolin~L. Xin.
\newblock Long-life lithium-ion batteries realized by low-ni, co-free cathode chemistry.
\newblock \emph{Nature Energy}, 8\penalty0 (7), 2023.

\bibitem[Zhang et~al.(2024)Zhang, Fu, Zhang, and Yan]{CJE4}
Tao Zhang, Ying Fu, Jun Zhang, and Chenggang Yan.
\newblock Deep guided attention network for joint denoising and demosaicing in real image.
\newblock \emph{Chinese Journal of Electronics}, 33\penalty0 (1):\penalty0 303--312, 2024.

\bibitem[Zhou et~al.(2024)Zhou, Min, Liu, Jin, Yu, and Zhang]{functional2}
Xinfeng Zhou, Peng Min, Yue Liu, Meng Jin, Zhong-Zhen Yu, and Hao-Bin Zhang.
\newblock Insulating electromagnetic-shielding silicone compound enables direct potting electronics.
\newblock \emph{Science}, 385\penalty0 (6714):\penalty0 1205--1210, 2024.

\end{thebibliography}
}


\end{document}